\documentclass{llncs}

\usepackage{cite}
\usepackage{times}
\usepackage{epsfig}
\usepackage{graphicx}
\usepackage{amsmath}
\usepackage{amssymb}
\usepackage{multirow}
\usepackage{makecell}
\usepackage{adjustbox}
\usepackage{tabularx}
\usepackage{placeins}
\usepackage{textcomp}
\usepackage{float}

\usepackage[pagebackref=true,breaklinks=true,letterpaper=true,colorlinks,bookmarks=false]{hyperref}

\newcommand{\repeatthanks}{\textsuperscript{\thefootnote}}

\begin{document}

\title{Grader variability and the importance of reference standards for evaluating machine learning models for diabetic retinopathy
}

\author{
  Jonathan Krause\textsuperscript{1}, Varun Gulshan\textsuperscript{1}, Ehsan Rahimy\textsuperscript{2}, Peter Karth\textsuperscript{3}, Kasumi Widner\textsuperscript{1}, Greg S. Corrado\textsuperscript{1}, Lily Peng\textsuperscript{1}\thanks{Equal contribution}, Dale R. Webster\textsuperscript{1}\repeatthanks \\
  \quad\\
  \textsuperscript{1}Google Research \qquad \textsuperscript{2}Palo Alto Medical Foundation \qquad \textsuperscript{3}Oregon Eye Consultants\\
}

\institute{}

\maketitle

\begin{abstract}
Diabetic retinopathy (DR) and diabetic macular edema are common complications of diabetes which can lead to vision loss. The grading of DR is a fairly complex process that requires the detection of fine features such as microaneurysms, intraretinal hemorrhages, and intraretinal microvascular abnormalities. Because of this, there can be a fair amount of grader variability. There are different methods of obtaining the reference standard and resolving disagreements between graders, and while it is usually accepted that adjudication until full consensus will yield the best reference standard, the difference between various methods of resolving disagreements has not been examined extensively. In this study, we examine the variability in different methods of grading, definitions of reference standards, and their effects on building deep learning models for the detection of diabetic eye disease. We find that a small set of adjudicated DR grades allows substantial improvements in algorithm performance. The resulting algorithm's performance was on par with that of individual U.S. board-certified ophthalmologists and retinal specialists.
\end{abstract}

\section{Introduction}
Diabetic retinopathy (DR) and diabetic macular edema (DME) are among the leading causes of vision loss worldwide. Several methods for assessing the severity of diabetic eye disease have been established, including the Early Treatment Diabetic Retinopathy Study Grading System~\cite{early1991grading}, the Scottish Diabetic Retinopathy Grading System~\cite{scotland}, and the International Clinical Diabetic Retinopathy (ICDR) disease severity scale~\cite{icdr}. The ICDR scale is one of the more commonly used clinical scales and consists of a 5-point grade for DR: no, mild, moderate, severe, and proliferative.

The grading of DR is a fairly complex process that requires the identification and quantification of fine features such as microaneurysms (MAs), intraretinal hemorrhages, intraretinal microvascular abnormalities, and neovascularization.  As a result, there can be a fair amount of grader variability, with intergrader kappa scores ranging from 0.40 to 0.65~\cite{early1991grading,scott2008agreement,li2010digital,gangaputra2013comparison,ruamviboonsuk2006interobserver}. This is not surprising because grader variability is a well-known issue with human interpretation of imaging in other medical fields such as radiology~\cite{elmore1994variability} or pathology~\cite{elmore2015diagnostic}.

Several methods have been proposed for resolving disagreements between graders and obtaining a reference standard. One approach consists of taking the majority decision from a group of 3 or more independent graders.  Another consists of having a group of 2 or more graders work independently and then having a third generally more senior grader arbitrate disagreements, with that individual's decision serving as the reference standard. Last, a group of 3 or more graders may first grade independently and then collectively discuss disagreements until there is full consensus on the final grade. The difference between various methods of resolving disagreements has not been examined extensively.

Deep learning~\cite{lecun2015deep} is a family of machine learning techniques that allows computers to learn the most predictive features directly from images, given a large dataset of labeled examples, without specifying rules or features explicitly. It has also been applied recently in medical imaging, producing highly accurate algorithms for a variety of classification tasks, including melanoma~\cite{esteva2017dermatologist}, breast cancer lymph node metastasis~\cite{liu2017detecting,bejnordi2017diagnostic}, and DR~\cite{gulshan2016development,gargeya2017automated,ting2017development} Because the network is trained to predict labels that have been paired with the images, it is imperative that the labels accurately represent the state of disease found in the image, especially for the evaluation sets.

In this study, we examine the variability in different methods of grading, definitions of reference standards, and their effects on building deep learning models for the detection of diabetic eye disease.

\section{Methods}
\subsection{Development Dataset}
In this work, we built upon the datasets used by Gulshan et al.~\cite{gulshan2016development} for algorithm development and clinical validation. A summary of the various data sets and grading protocols used for this study is shown in Figure~\ref{fig:image_source_protocols}. The development dataset used consists of images obtained from patients who presented for DR screening at EyePACS-affiliated clinics, 3 eye hospitals in India (Aravind Eye Hospital, Sankara Nethralaya, and Narayana Nethralaya), and the publicly available Messidor-2 dataset~\cite{decenciere2014feedback,quellec2008optimal}\footnote{We have released our adjudicated grades on the Messidor-2 dataset at \url{https://www.kaggle.com/google-brain/messidor2-dr-grades}.}. Images from the EyePACS clinics consisted of 45\textdegree\ retinal fundus images that were primary (or posterior pole-centered), nasal (disc-centered), and temporal field of view. The rest of the data sources only had primary field of view images. The images were captured using the Centervue DRS, Optovue iCam, Canon CR1/DGi/CR2, and Topcon NW using 45\textdegree\ fields of view. All images were de-identified according to Health Insurance Portability and Accountability Act Safe Harbor before transfer to study investigators.  Ethics review and institutional review board exemption was obtained using Quorum Review Institutional Review Board.

There were 3 types of grades used in the development dataset: grading by EyePACS graders, grading by ophthalmologists, and adjudicated consensus grading by retinal specialists. Images were provided at full resolution for manual grading. The images provided by EyePACS clinics were graded according to the EyePACS protocol~\cite{eyepacs}. In the EyePACS protocol, all 3 images of an eye (primary, nasal, and temporal) were graded together and assigned a single grade. We assigned the grade provided for the entire eye as a grade for each image. Although the eye-level grade may not correspond exactly with the image-level grade, this method allowed for training on different fields of view, which helps the algorithm tolerate some deviation from the primary field of view.  The second source of grades was the development dataset used by Gulshan et al.~\cite{gulshan2016development}, where each image was independently graded by ophthalmologists or trainees in their last year of residency. The third type of grades was obtained using our subspecialist grading and adjudication protocol on a small subset of images. Our subspecialist adjudication protocol consisted of 3 fellowship-trained retina specialists undertaking independent grading of the dataset. Next, each of the 3 retina specialists reviewed the images with any level of disagreement in a combination of asynchronous and live adjudication sessions.  

For algorithm development, the development dataset was divided into groups: a ``train'' set and a ``tune'' set. The ``train'' set consisted of all the images that were not adjudicated and was used to train the model parameters. The ``tune'' set consisted of images with adjudicated consensus grades and was used to tune the algorithm hyperparameters (e.g., input image resolution, learning rate) and make other modeling choices (e.g., network architectures). All of the images in the ``tune'' set were of the primary field.

\subsection{Clinical Validation Dataset}
The clinical validation dataset consisted of primary field of view images that were acquired from EyePACS clinics (imaged between May and October 2015). These did not overlap with any images or patients used in the development dataset.

The clinical validation set was graded using the same adjudication protocol as the tuning set (face-to-face adjudication by 3 retina specialists). In addition, the clinical validation set was graded by 3 ophthalmologists. These ophthalmologists were distinct from the retina specialists who adjudicated the set. After grading, all disagreements between the adjudicated consensus of retinal specialists and majority decision of the ophthalmologists were reviewed manually by a retinal specialist, who also assigned a likely reason for the discrepancy.

\subsection{Algorithm Training}
Our deep learning algorithm for predicting DR and DME was built upon the architecture used by Gulshan et al.~\cite{gulshan2016development}. We used a convolutional neural network~\cite{lecun1998gradient} that predicted a 5-point DR grade, referable DME, and gradability of image. The input to the neural network was an image of a fundus, and through the use of many stages of computation, parameterized by millions of numbers, the network output a real-valued number between 0 and 1 for each prediction, indicating its confidence.

The parameters of a neural network were determined by training it on a dataset of fundus images. Repeatedly, the model was given an image with a known severity rating for DR, and the model predicted its confidence in each severity level of DR, slowly adjusting its parameters over the course of the training process to become more accurate. The model was trained via distributed stochastic gradient descent and evaluated on a tuning dataset throughout the training process. The tuning dataset was used to determine model hyperparameters (parameters of the model architecture that cannot be trained with gradient descent). Finally, we created an ``ensemble'' of models, training 10 individual models and combining their predictions, which improves performance and robustness.

\subsection{Algorithmic Improvements}
We made a number of improvements to the core neural network, compared with the study by Gulshan et al.~\cite{gulshan2016development} First, we trained our model on a larger set of images obtained from EyePACS~\cite{cuadros2009eyepacs}. Because it would be prohibitively expensive to relabel all of these images with U.S.-licensed ophthalmologists (as done by Gulshan et al.~\cite{gulshan2016development}, we also use labels determined by the EyePACS grading centers to train our DR classifier. To use both our ophthalmologist grading and the EyePACS grading effectively, we treated the labels from these 2 independent sources as separate prediction targets, training our model to predict both an EyePACS grade for DR and a grade determined by our own labeling procedure, if present.

By using this larger training set, we performed a more extensive search for well-performing hyperparameters using a Gaussian process bandit algorithm~\cite{golovin2017google} One significant change in model hyperparameters that resulted was an increase in the resolution of images the model used as input: The model used in this work has an input resolution of 779 x 779 pixels, a large increase over the 299 x 299 pixels used by Gulshan et al.~\cite{gulshan2016development}. This has the effect of increasing the effective resolution of lesions. For example, an MA or small hard exudate may occupy only 1 pixel or less within a 299 x 299 pixel image~\cite{kolb1995facts,scanlon2017practical}, but at this higher resolution, it would occupy approximately 6.8 pixels. The effective resolution of other lesions is similarly increased when using higher input resolution.

Other significant changes included changing model architecture from ``Inception-v3''~\cite{szegedy2016rethinking} to ``Inception-v4''~\cite{szegedy2017inception} and predicting a 5-class rating for DR (Gulshan et al.~\cite{gulshan2016development} only predicted binary referables). For other hyperparameter and training details, see Appendix Section~\ref{sec:hyperparameters}.

\subsection{Evaluation}
One advantage of using machine learning-based systems for predicting DR is that they explicitly output a confidence in the range (0e1) for each image. By applying a threshold to this confidence to determine referability, an algorithm's sensitivity and specificity can be tuned for a particular use case; as the threshold is lowered, the algorithm becomes more sensitive, but less specific. Performance across many different thresholds is depicted in a receiver operating curve and is summarized with the area under the curve (AUC). We used as our primary model evaluation metric for DR the AUC at different DR severity thresholds, for example, the AUC for predicting moderate and above DR. In addition, for some experiments on 5-class prediction we evaluated using the quadratic-weighted Cohen's kappa~\cite{cohen1960coefficient}.

The algorithm's sensitivity and specificity when predicting a 5-class output cannot be tuned with a single threshold, though.  Therefore, we used a cascade of thresholds, building on the model's predicted confidence in each of the severity levels for DR. First, a threshold was picked for proliferative DR: If the model's predicted probability of an image corresponding to proliferative DR was above a specified threshold, chosen to reach a target sensitivity on the tuning set, the image was classified as proliferative DR. Next, another threshold was picked for the model's predicted probability of severe DR and applied to all images not classified as proliferative DR. A similar procedure was applied for both moderate and mild DR. Finally, all remaining images whose probabilities did not pass any of the thresholds were classified as no DR.

\section{Results}
\subsection{Grading and Adjudication}
The baseline characteristics of the development and clinical validation datasets are described in Table~\ref{table:baseline_characteristics}. The training portion of the development set consisted of more than 1.6 M fundus images from 238 610 unique individuals. The tune portion of the development set consisted of adjudicated consensus grades for 3737 images from 2643 unique individuals. The clinical validation set consisted of 1958 photos from 998 unique individuals. Compared with the clinical validation set, the development set had a slightly higher proportion of abnormal images.

A comparison of the grades generated by the majority decision from independent grading by the retinal specialists in the panel before the adjudication process versus the adjudicated reference standard is outlined in Table~\ref{table:comparison_retinal_specialist}. Most grades remained within 1 step of each level of severity. For DR, there were 27 instances (1.5\% of images) where the difference was 2 steps, and none that were 3 or more. Adjudication more often yielded new grades with higher levels of severity. The weighted-kappa score for the grade determined by the majority decision of retinal specialists and by adjudication was 0.91 (Table~\ref{table:agreement_specialist}).

To better characterize the difference in each retinal specialist's preadjudication grade and the adjudicated consensus for detecting moderate and above DR, we measured the sensitivity and specificity of each specialist (Table~\ref{table:agreement_specialist}). Although all 3 specialists were very specific (~99\%), sensitivity ranged from 74.4\% to 82.1\%, corresponding to weighted-kappa scores of 0.82 to 0.91. A calculation of the nonadjudicated majority decision of the panel of 3 retina specialists showed higher sensitivity (88.1\%) than grades from individual specialists (Table 3), leaving 11.9\% of further cases with moderate or above DR that required the adjudication process to detect. During independent grading, some retinal specialists used a different variation for grading referable DME in which they considered any hard exudate within the macular OCT box (~2 disc diameters) as referable. This discrepancy was resolved during adjudication, and all images where referable DME was potentially present were reassessed to match the standard protocol (where hard exudates within 1 disc diameter was considered referable DME). For this reason, the independent grading of the retinal specialists for DME was not included in subsequent analyses.

In addition to having retinal specialists grade and adjudicate the clinical validation datasets, we also had 3 U.S. board-certified ophthalmologists grade the same set (Table~\ref{table:comparison_oph}). Quadratic-weighted kappa values were generally good (0.80 - 0.84), but somewhat lower for ophthalmologists than for the retina specialists (Table~\ref{table:agreement_oph}). The majority decision of the ophthalmologists yielded a higher agreement (weighted kappa: 0.87) than individual ophthalmologists alone. A comparison of all graders (3 retinal specialists and 3 ophthalmologists) showed that disagreements were more common in cases in which the adjudicated consensus yielded referable disease (Figure~\ref{fig:grader_agreement}). An analysis combining the 3 retinal specialists and 3 ophthalmologists is given in Table~\ref{table:comparison_specialist_and_oph}.

A summary of the causes of errors in the nonadjudicated grades generated from the majority decision of the ophthalmologists is presented in Table~\ref{table:reasons}. The most common causes were missed MAs (36\%), artifacts (20\%), and disagreement regarding whether a lesion was an MA or a hemorrhage (16\%). Thirty-six of 193 disagreements differed by at least 2 grades (e.g., no DR vs.  moderate or moderate vs. proliferative). Of these, overgrading, cases in which the majority decision of the ophthalmologists was more severe than the final adjudicated consensus grade, accounted for 14 (7.3\%) of the disagreements, and undergrading accounted for 22 (11.4\%) of the cases.

\subsection{Model Results}
We compare the results of our model with the model used by Gulshan et al.~\cite{gulshan2016development} in Table~\ref{table:model_aucs}, using the majority decision or adjudicated consensus of retinal specialist grades. For moderate or worse DR, we observed an improvement in AUC with increasing image resolution as well as with the other algorithmic improvements outlined in the ``Methods'' section for both methods of computing reference standard. However, this improvement in AUC was larger when using the adjudicated consensus as the reference standard (from 0.942 to 0.986 AUC). When predicting mild or worse DR (Fig~\ref{fig:res_auc}), increasing image resolution showed only a slight gain (<1\%) when using majority decision as the reference standard, while when evaluating with adjudicated consensus grades we observed that increasing resolution improved the model's AUC from 0.952 to 0.984. However, increasing resolution beyond approximately 450 x 450 pixel input seemed to have marginal performance gains. Another benefit of evaluating with adjudicated consensus grades was in reduced metric variability, where we observed reduced metric variance (confidence intervals) across input resolutions. This was particularly important when making changes to the algorithm because it made it easier to determine with confidence if a change in the algorithm was actually an improvement.

We illustrate the performance of our final model using the adjudicated consensus as the reference standard in Table~\ref{table:comparison_alg}. On 5-class DR prediction, our model achieved a quadratic-weighted kappa of 0.84, which is on par with the performance of individual ophthalmologists and individual retinal specialists. There were 257 disagreements between the adjudicated consensus and the model. Of the 56 (21.8\%) disagreements that were 2 or more steps, 53 (20.6\%) were the result of overgrading, that is, the model predicted a higher grade than the adjudicated consensus, and 3 (1.2\%) resulted from undergrading.

In addition to 5-point grading analysis for DR, we also demonstrate the model's performance in binary classification tasks.  Figure~\ref{fig:roc_curves} summarizes the model's performance as well as that of the ophthalmologists and retinal specialists in detecting mild or worse DR, moderate or worse DR, severe or worse DR, and proliferative DR. For all of these classification tasks, the algorithm's performance was approximately on par with that of individual ophthalmologists and retinal specialists. For mild or worse DR, the algorithm had a sensitivity of 0.970, specificity of 0.917, and AUC of 0.986.

\section{Discussion}
Deep learning has garnered attention recently because of its ability to create highly accurate algorithms from large datasets of labeled data without feature engineering. However, great care must be taken to evaluate these algorithms against as high-quality ``ground truth'' labels as possible.  Previous work~\cite{gulshan2016development} used majority decision to generate ground truth, without an adjudication process. The present work suggests that an adjudication process yielding a consensus grade from multiple retina specialists provides a more rigorous reference standard for algorithm development, especially for the identification of artifacts and missed MAs, ultimately leading to superior algorithms and improved detection of disease.

The adjudication sessions of the 3 retina specialists yielded considerable insights into the grading process.  First, image artifacts, especially those resembling typical pathology such as MAs, were a common source of disagreement. These disagreements were consistent across many images, because artifacts are often caused by consistent lens artifacts or dust spots. The adjudication process was critical in correcting these errors. Likewise, when true pathology was present, a frequent cause of grader disagreement was in differentiating an MA from an intraretinal hemorrhage, which in the absence of other findings, represented the difference between labeling an image as mild or moderate DR. In addition, there was significant subjective variance in the precise definition of the boundaries between grades at the outset of the adjudication process. Although traditional grading definitions are fixed, it became very clear that there are gray areas where these definitions are not sufficient to make an unambiguous call on DR severity, for example, on whether a macula-centered image with numerous hemorrhages was more likely to represent moderate or severe DR. Once all 3 retina specialists converged on a more precise set of definitions, staying within but also refining the traditional definitions, many of these initial disagreements were resolved. On conclusion of the adjudication process, each of the retina specialists indicated that the precision used in the adjudication process was above that typically used in everyday clinical practice.

A comparison of the performance of algorithm and manual grading also yielded interesting insights. First, although manual grading by ophthalmologists led to an approximately even ratio of overgrading (7.3\%) and undergrading (11.4\%), the grading by the algorithm had a higher ratio of overgrading (20.36\%) than undergrading (1.2\%). This can be attributed to the decision to operate the algorithm at a very high-sensitivity set point, which would be desirable for a screening algorithm to maximize the detection of referable DR. However, a combination of the algorithm for initial screening coupled with manual review of the resulting positive images would likely yield a system that is both highly sensitive and specific, which may help limit the number of diabetic patients being referred who do not actually have referable disease.

It is also worth noting that although full-resolution images were used for grading and adjudication, training models using an input resolution greater than 450 x 450 pixels did not yield a significant boost in model performance.  This suggests neural networks may not require as high of a resolution input image to detect DR lesions as humans do. However, this is an observation made in the setting of a specific grading scale (ICDR) using only single nonstereoscopic 45\textdegree\ color fundus photographs. This observation may not generalize if other finer grading scales are used, such as Early Treatment Diabetic Retinopathy Study, if the input images are different (stereoscopic, ultrawide-field), or if the ground truth incorporated signals from other clinical data, such as other types of retinal imaging like OCT or fluorescein angiography.

There are several limitations to this study that may represent further avenues for research. One key limitation is that we used hard exudates as a proxy for DME, which generally requires stereoscopic images or OCT to properly diagnose and characterize. Including OCT imaging in the process of determining ground truth for DME will be valuable in future studies.

In addition, we used only 3 retinal specialists to establish ground truth and 3 U.S. Board-Certified ophthalmologists as a comparison group. However, the kappa scores, sensitivity, and specificity of the graders in this study are in line with, if not slightly better than, what has been reported in other studies~\cite{gangaputra2013comparison,ruamviboonsuk2006interobserver,gonzalez1995concordance,abramoff2013automated} but we found a wider range for sensitivity, especially when it comes to detecting mild disease. For example, using adjudication as the ground truth, the 3 retinal specialists in Abr\`amoff et al.~\cite{abramoff2013automated} had preadjudication sensitivities of 71\% to 91\% and specificities of 95\% to 100\% for calling referable DR (defined as moderate or worse DR or referable DME). Likewise, our retinal specialists had preadjudication sensitivities of 74\% to 82\% and specificities of more than 99\% for moderate or worse DR. Gangaputra et al.~\cite{gangaputra2013comparison} reported sensitivities of 53\% and 84\% and specificities of 97\% and 96\% for clinical grading versus reading center grading for detecting mild or worse DR. The retinal specialists and ophthalmologists in our study had sensitivities between 70\% and 80\% and specificities more than 97\%. Examining the differences between adjudication and reading center grading or between different adjudication groups would be valuable next steps in investigating variability in grading. We also did not study the variability in grading for optometrists or other graders trained to evaluate DR images, which is another area for further exploration.

Other areas of future work include measuring the impact of using adjudicated consensus grades in the training portion of the development dataset on model performance and studying the algorithm's performance on other clinical validation datasets. Although images in the development and clinical validation datasets of our study were taken in nonoverlapping time intervals from the EyePACS population, additional validation on more diverse datasets in terms of cameras, geographies, and patient population would further characterize the algorithm's generalizability.

Despite these limitations, we believe this work provide a basis for further research and begins to lay the groundwork for standards in this field. For example, although adjudication yields a more reliable ground truth, it requires significant time and resources to perform. We demonstrate that by using existing grades and adjudicating a small subset (0.22\%) of the training image grades for the ``tune'' set, we were able to significantly improve model performance without adjudicating the entire training corpus of images.  Leveraging these techniques, our resulting model's performance was approximately on par with that of individual ophthalmologists and retinal specialists.

In conclusion, we believe that proper determination of ground truth is a critical aspect of the development of clinically useful machine learning algorithms for use in detection of or screening for retinal disease. We have shown that our process of live adjudication by multiple subspecialists yielding a consensus grade provides improved algorithmic accuracy and led to improved screening models.

\section{Acknowledgements}
From Google Research: Yun Liu, Derek Wu, Katy Blumer, Philip Nelson

\noindent
From EyePACS: Jorge Cuadros

\bibliographystyle{unsrt}
\bibliography{paper}

\appendix

\FloatBarrier
\setcounter{table}{0}
\renewcommand{\thetable}{\arabic{table}}%
\setcounter{figure}{0}
\renewcommand{\thefigure}{\arabic{figure}}%

\newpage
\section{Model Hyperparameters}
\label{sec:hyperparameters}
For our implementation of Gulshan et al.~\cite{gulshan2016development}, we used the following hyperparameters:
\begin{itemize}
  \item Inception-v3 architecture. This is very slightly different from Gulshan et al., who used an architecture just pre-dating Inception-v3.
  \item Input image resolution: $299 \times 299$
  \item Learning rate: 0.001
  \item Batch size: 32
  \item Weight decay: $4 \cdot 10^{-5}$
  \item An Adam optimizer with $\beta_1 = 0.9$, $\beta_2 = 0.999$, and $\epsilon = 0.1$. This is a difference from Gulshan et al., who used RMSProp.
  \item Data augmentation:
    \begin{itemize}
      \item Random horizontal reflections
      \item Random brightness changes (with a max delta of 0.125) [see the TensorFlow function \verb|tf.image.random_brightness|]
      \item Random saturation changes between 0.5 and 1.5 [see \verb|tf.image.random_saturation|]
      \item Random hue changes between -0.2 and 0.2 [see \verb|tf.image.random_hue|]
      \item Random contrast changes between 0.5 and 1.5 [see \verb|tf.image.random_constrast|]
      \item The above is done in that order. Note that Gulshan et al. applied the random brightness, saturation, hue, and contrast changes in a random order.
    \end{itemize}
  \item Early stopping based on tuning set DR Moderate AUC, which happened within the first 60,000 steps.
  \item Model evaluations performed using a running average of parameters, with a decay factor of 0.9999.
\end{itemize}

\noindent
In our full model we used the following hyperparameters:
\begin{itemize}
  \item Inception-v4 architecture
  \item Input image resolution: $779 \times 779$
  \item Learning rate: 0.0014339
  \item Batch size: 24
  \item Weight decay: $1.15 \cdot 10^{-5}$
  \item An Adam optimizer with $\beta_1 = 0.9$, $\beta_2 = 0.999$, and $\epsilon = 0.1$.
  \item Data augmentation:
    \begin{itemize}
      \item Random vertical and horizontal reflections
      \item Random brightness changes (with a max delta of 0.5247078)
      \item Random saturation changes between 0.3824261 and 1.4029386
      \item Random hue changes between -0.1267652 and 0.1267652
      \item Random contrast changes between 0.3493415 and 1.3461331
      \item The above is done in that order.
    \end{itemize}
  \item Each model in a 10-way ensembled was trained for 250,000 steps.
  \item Model evaluations performed using a running average of parameters, with a decay factor of 0.9999.
\end{itemize}

\newpage
\section*{Tables and Figures}

\begin{table}[h!]
  \centering
\resizebox{\columnwidth}{!}{%
  \begin{tabular}{|l|c|c|c|}
    \hline
    & \multicolumn{2}{c|}{\textbf{Development}} & \textbf{Clinical Validation} \\ \hline
    & Train & Tune & EyePACS-2 \\ \hline
    Images (\#) & 1,665,151 & 3,737 & 1,958 \\ \hline
    \multicolumn{4}{|l|}{\textbf{Patient Demographics}} \\ \hline
    Unique Individuals (\#) & 238,610 & 2,643 & 999 \\ \hline
    Age (Average $\pm$ Stdev) & 53.5 $\pm$ 11.6 & 54.3 $\pm$ 11.1 & 54.9 $\pm$ 10.9 \\ \hline
    \makecell[l]{Female / Total patients \\ where gender was known} & 140,183/230,556 (60.8\%) & 1,101/1,767 (62.3\%) & 603/999 (60.5\%) \\ \hline
    \multicolumn{4}{|l|}{\textbf{Image Quality Distribution}} \\ \hline
    \makecell[l]{Fully gradable / Total images \\ where image quality was \\ assessed} & 1,343,726/1,529,771 (87.8\%) & 3,547/3,737 (94.9\%) & 1,813/1,958 (92.6\%) \\ \hline
  \end{tabular}
}
\vspace{5mm}
\newline
  \begin{tabular}{|l|c|c|c|c|c|c|}
    \hline
    & \multicolumn{4}{c|}{\textbf{Development}} & \multicolumn{2}{c|}{\textbf{Validation}} \\ \hline
    & \multicolumn{2}{c|}{Train} & \multicolumn{2}{c|}{Tune} & \multicolumn{2}{c|}{EyePACS-2} \\ \hline
    \makecell[l]{\textbf{Disease Severity}\\\textbf{Distribution}} & \# & \% & \# & \% & \# & \% \\ \hline
    \makecell[l]{Total images where DR \\ was assessed} & \, 1,662,646 \, & \, 100.0\% \, & \, 3,547 \, & \, 100.0\% \, & \, 1,813 \, & \, 100.0\% \, \\ \hline
    No diabetic retinopathy & 1,164,368 & 70.0\% & 2,417 & 68.1\% & 1,478 & 81.5\% \\ \hline
    Mild & 152,938 & 9.2\% & 458 & 12.9\% & 125 & 6.9\% \\ \hline
    Moderate & 249,138 & 15.0\% & 497 & 14.0\% & 144 & 7.9\% \\ \hline
    Severe & 50,640 & 3.0\% & 106 & 3.0\% & 50 & 2.8\% \\ \hline
    Proliferative & 45,562 & 2.7\% & 69 & 1.9\% & 16 & 0.9\% \\ \hline
    \multicolumn{7}{|c|}{} \\ \hline
    \makecell[l]{Total images where DME \\ was assessed} & 252,544 & 100.0\% & 3,547 & 100.0\% & 1,813 & 100.0\% \\ \hline
    \makecell[l]{Referable diabetic macular \\ edema} & 28,807 & 11.4\% & 281 & 7.9\% & 79 & 4.4\% \\ \hline
  \end{tabular}
  \vspace{4mm}
  \caption{
    Summary of image characteristics and available demographic information in the development and clinical validation datasets. The adjudicated reference standard was used for computing the DR and DME distributions on the ``Tune'' and ``EyePACS-2'' datasets, and the majority reference standard was used for the ``Train'' dataset. For DR, the majority was taken over both the EyePACS partner grade, and the grading done by our ophthalmologists (using whichever grades are available). For DME, the majority is taken over grades from our ophthalmologists. For Image quality distributions, the EyePACS partner grade was used for the ``Train'' dataset and an adjudicated image quality standard was used for the ``Tune'' and ``EyePACS-2'' datasets. The ``Tune'' dataset and ``Validation'' dataset differ in their definition of DME used during adjudication: the ``Tune'' dataset used a standard of hard exudates within 2 disc diameter (DD) of the fovea (to approximate the macular OCT box), whereas the ``Validation'' dataset used a standard of hard exudates within 1 DD of the fovea.
  }
  \label{table:baseline_characteristics}
\end{table}
\newpage

\begin{table}[t]
  \centering
  \begin{tabular}{|l|l|r|r|r|r|r|}
    \hline
    & & \multicolumn{5}{c|}{\makecell{\textbf{Majority decision of retinal specialist grading}\\\textbf{before adjudication}}} \\ \hline
    & & \textbf{No} & \, \textbf{Mild} & \, \textbf{Moderate} \, & \textbf{Severe} & \, \textbf{Proliferative} \\ \hline
    \multirow{ 5}{*}{\makecell{\textbf{Adjudicated}\\\textbf{Consensus}}} & \textbf{No} & \, 1,469 & 4 & 5 & 0 & 0 \\ \cline{2-7}
    & \textbf{Mild} & 58 & 62 & 5 & 0 & 0 \\ \cline{2-7}
    & \textbf{Moderate} & 22 & 3 & 118 & 1 & 0 \\ \cline{2-7}
    & \textbf{Severe} & 0 & 0 & 13 & 36 & 1 \\ \cline{2-7}
    & \textbf{Proliferative} & 0 & 0 & 0 & 1 & 15 \\ \hline
  \end{tabular}
  \vspace{4mm}
  \caption{
Comparison of retinal specialist grades before and after adjudication on the validation dataset. Confusion matrix for diabetic retinopathy between the grade determined by majority decision and adjudicated consensus.
  }
  \label{table:comparison_retinal_specialist}
  \bigbreak
  \bigbreak
  \centering
\resizebox{0.75\columnwidth}{!}{%
  \begin{tabular}{|c|c|c|c|c|c|c|}
    \hline
    \multirow{2}{*}{} & \multicolumn{3}{c|}{\textbf{DR}} \\ \cline{2-4}
    & \, Sensitivity \, & \, Specificity \, & \, \makecell{Quadratic- \\ weighted kappa} \, \\ \hline
    \, Retina Specialist A \, & 74.6\% & 99.1\% & 0.82  \\ \hline
    Retina Specialist B & 74.4\% & 99.3\% & 0.80 \\ \hline
    Retina Specialist C & 82.1\% & 99.3\% & 0.91 \\ \hline
    \makecell{Majority Decision \\ (Retinal Specialists)} & 88.1\% & 99.4\% & 0.91  \\ \hline
  \end{tabular}
}
  \vspace{4mm}
  \caption{
Agreement between each retina specialist and the adjudicated reference standard on the validation dataset.
Retina specialists correspond to those who contributed to the final adjudicated reference standard. Sensitivity and specificity metrics reported are for moderate or worse DR. Agreement between the preadjudication 5-point DR grade and the final adjudicated grade is also measured by the quadratic-weighted kappa.
  }
  \label{table:agreement_specialist}
\end{table}

\begin{table}[hp]
  \centering
  \begin{tabular}{|l|l|r|r|r|r|r|}
    \hline
    & & \multicolumn{5}{c|}{\makecell{\textbf{Majority of ophthalmologist grading}}} \\ \hline
    & & \textbf{No} & \, \textbf{Mild} & \, \textbf{Moderate} \, & \textbf{Severe} & \, \textbf{Proliferative} \\ \hline
    \multirow{ 5}{*}{\makecell{\textbf{Adjudicated}\\\textbf{Consensus}}} & \textbf{No} & \, 1,432 & 32 & 10 & 0 & 1 \\ \cline{2-7}
    & \textbf{Mild} & 67 & 38 & 19 & 1 & 0 \\ \cline{2-7}
    & \textbf{Moderate} & 20 & 14 & 94 & 14 & 2 \\ \cline{2-7}
    & \textbf{Severe} & 0 & 0 & 9 & 41 & 0 \\ \cline{2-7}
    & \textbf{Proliferative} & 0 & 0 & 1 & 2 & 13 \\ \hline
  \end{tabular}
\vspace{5mm}
\newline
  \begin{tabular}{|l|l|r|r|}
    \hline
    & & \multicolumn{2}{c|}{\makecell{\textbf{Majority of ophthalmologist}\\ \textbf{grading}}} \\ \hline
    & & \, \textbf{Not Referable} & \, \textbf{Referable DME} \\ \hline
    \multirow{ 2}{*}{\makecell{\textbf{Adjudicated}\\\textbf{Consensus}}} & \textbf{Not Referable} & 1,710 & 17 \\ \cline{2-4}
    & \textbf{Referable DME} & 13 & 65 \\ \hline
  \end{tabular}
  \vspace{4mm}
  \caption{
Comparison of ophthalmologist grades versus adjudicated grades from retina specialists on the validation dataset. Confusion matrix for diabetic retinopathy and DME between the grade determined by majority decision of the ophthalmologists and the adjudicated consensus of retinal specialists.
  }
  \label{table:comparison_oph}

  \bigbreak
  \bigbreak

  \centering
\resizebox{\columnwidth}{!}{%
  \begin{tabular}{|c|c|c|c|c|c|c|}
    \hline
    \multirow{2}{*}{} & \multicolumn{3}{c|}{\textbf{DR}} & \multicolumn{2}{c|}{\textbf{DME}} \\ \cline{2-6}
    & \, Sensitivity \, & \, Specificity \, & \, \makecell{Quadratic- \\ weighted kappa} \, & \, Sensitivity \, & \, Specificity \, \\ \hline
    \, Ophthalmologist A \, & 75.2\% & 99.1\% & 0.84 & 81.5\% & 98.7\% \\ \hline
    Ophthalmologist B & 74.9\% & 97.9\% & 0.80 & 62.7\% & 98.6\% \\ \hline
    Ophthalmologist C & 76.4\% & 97.5\% & 0.82 & 86.4\% & 99.1\% \\ \hline
    \makecell{Majority Decision \\ (Ophthalmologists)} & 83.8\% & 98.1\% & 0.87 & 83.3\% & 99.0\% \\ \hline
  \end{tabular}
}
  \vspace{4mm}
  \caption{
Agreement between ophthalmologists' grades with the adjudicated reference standard on the validation dataset. Sensitivity and specificity metrics are for moderate or worse DR and referable DME for each grader. Agreement between the adjudicated grade and the 5-point scale is also measured by the quadratic-weighted kappa.
  }
  \label{table:agreement_oph}
\end{table}

\begin{table}[t]
  \centering
  \begin{tabular}{|l|l|l|l|l|l|l|}
    \hline
    & \multicolumn{6}{c|}{\makecell{\textbf{Adjudication Consensus Grade (Retina Specialists) minus}\\\textbf{Majority Decision Grade (Ophthalmologists)}}} \\ \hline
    & \textbf{-4} \quad\quad\quad  & \textbf{-2} \quad\quad\quad  & \textbf{-1} \quad\quad\quad  & \textbf{1} \quad\quad\quad  & \textbf{2} \quad\quad\quad   & \textbf{Total} \quad\quad\quad   \\ \hline
    Artifact vs not & &5 &28 &1 &5 &39 \\ \hline
    Extent of Lesions & &1 &16 &9 & &26 \\ \hline
    Hemorrhage vs MA & &1 &13 &13 &3 &30 \\ \hline
    Hemorrhage vs not & &4 & &4 &11 &19 \\ \hline
    IRMA vs not & &1 & & & &1 \\ \hline
    Missed hemorrhage & & &2 & & &2 \\ \hline
    Missed MA & & & 6& 63& & 69\\ \hline
    Missed NVD/NVE & & & &2 &1 &3 \\ \hline
    PRP vs not & 1& 1& & & 1& 3\\ \hline
    Other & & & & &1 &1 \\ \hline
    Total & 1& 13& 65& 92& 22& 193\\ \hline
  \end{tabular}
  \vspace{4mm}
  \caption{
Reasons for difference between adjudication of retinal specialist and majority decision from ophthalmologist graders.  Disagreements between the adjudicated consensus and majority decision were examined and characterized by a retinal specialist. Positive numbers denote that the adjudication grade was more than the majority decision of ophthalmologist grade, and vice-versa for negative numbers.
  }
  \label{table:reasons}
\end{table}

\begin{table}[t]
  \centering
  \begin{tabular}{|l|c|c|c|}
    \hline
    & \textbf{Original Model} & \makecell{\textbf{Original Model,}\\\textbf{High Resolution}} & \textbf{Our Full Model} \\ \hline
    \makecell{\textbf{Majority decision of }\\\textbf{retinal specialists ground truth}} & 0.952 & 0.978 & 0.986 \\  \hline
    \makecell{\textbf{Adjudicated consensus}\\\textbf{ground truth}} & 0.942 & 0.960 & 0.986 \\  \hline
  \end{tabular}
  \vspace{4mm}
  \caption{
    Differences in the Final Area under the Curve Observed for Algorithms Trained in the Image Resolution Algorithm Selection Experiments for Moderate or Worse Diabetic Retinopathy. Majority decision was based on the grades of the 3 retinal specialists. ``Original Model'' refers to the model and training data of Gulshan et al.~\cite{gulshan2016development} ``Original Model, High Resolution'' uses higher resolution input images (779 x 779 pixels), and ``Our Full Model'' incorporates all changes described in the ``Algorithmic Improvements'' section in the main text.
  }
  \label{table:model_aucs}
\end{table}

\begin{table}[t]
  \centering
  \begin{tabular}{|l|l|r|r|r|r|r|}
    \hline
    & & \multicolumn{5}{c|}{\makecell{\textbf{Algorithm Grade}}} \\ \hline
    & & \textbf{No} & \, \textbf{Mild} & \, \textbf{Moderate} \, & \textbf{Severe} & \, \textbf{Proliferative} \\ \hline
    \multirow{ 5}{*}{\makecell{\textbf{Adjudicated}\\\textbf{Consensus}}} & \textbf{No} & \, 1,356 & 74 & 44 & 1 & 3 \\ \cline{2-7}
    & \textbf{Mild} & 7 & 43 & 74 & 0 & 1 \\ \cline{2-7}
    & \textbf{Moderate} & 3 & 3 & 98 & 36 & 4 \\ \cline{2-7}
    & \textbf{Severe} & 0 & 0 & 2 & 47 & 1 \\ \cline{2-7}
    & \textbf{Proliferative} & 0 & 0 & 0 & 4 & 12 \\ \hline
  \end{tabular}
\vspace{5mm}
\newline
  \begin{tabular}{|l|l|r|r|}
    \hline
    & & \multicolumn{2}{c|}{\makecell{\textbf{Algorithm Grade}}} \\ \hline
    & & \, \textbf{Not Referable} & \, \textbf{Referable DME} \\ \hline
    \multirow{ 2}{*}{\makecell{\textbf{Adjudicated}\\\textbf{Consensus}}} & \textbf{Not Referable} & 1,637 & 97 \\ \cline{2-4}
    & \textbf{Referable DME} & 4 & 75 \\ \hline
  \end{tabular}
\vspace{5mm}
\newline
  \centering
  \begin{tabular}{|c|c|c|c|c|c|}
    \hline
    \multirow{2}{*}{} & \multicolumn{3}{c|}{\textbf{DR}} & \multicolumn{2}{c|}{\textbf{DME}} \\ \cline{2-6}
    & \, Sensitivity \, & \, Specificity \, & \, \makecell{Quadratic- \\ weighted kappa} \, & \, Sensitivity \, & \, Specificity \, \\ \hline
    Algorithm & 97.1\% & 92.3\% & 0.84 & 94.9\% & 94.4\% \\ \hline
  \end{tabular}
  \vspace{4mm}
  \caption{
Comparison of algorithm grade versus adjudicated grades from retina specialists. Confusion matrix for DR and DME where the grade is determined by the algorithm or adjudicated consensus of the retinal specialists. Sensitivity and specificity metrics reported are for moderate or worse DR and referable DME.
}
  \label{table:comparison_alg}
\end{table}

\begin{figure*}[t]
  \centering
  \includegraphics[width=0.99\linewidth]{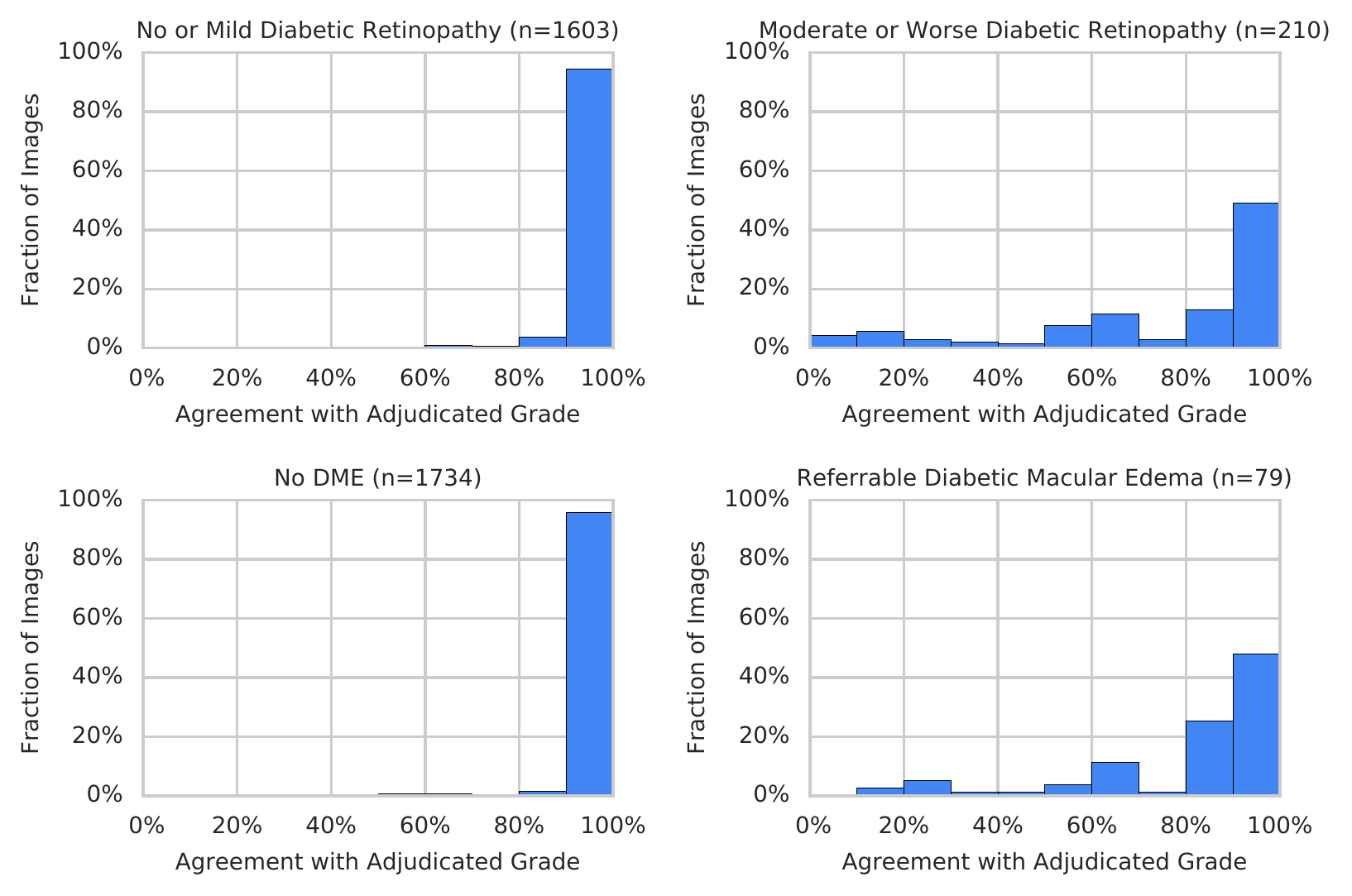}
  \caption{
Grader agreement based on the adjudicated consensus grade for referable diabetic retinopathy (DR) and diabetic macular edema (DME).  Independent grading of all 3 retinal specialists and all 3 ophthalmologists are included in this analysis.
  }
  \label{fig:grader_agreement}

  \bigbreak
  \bigbreak

  \centering
  \includegraphics[width=0.99\linewidth]{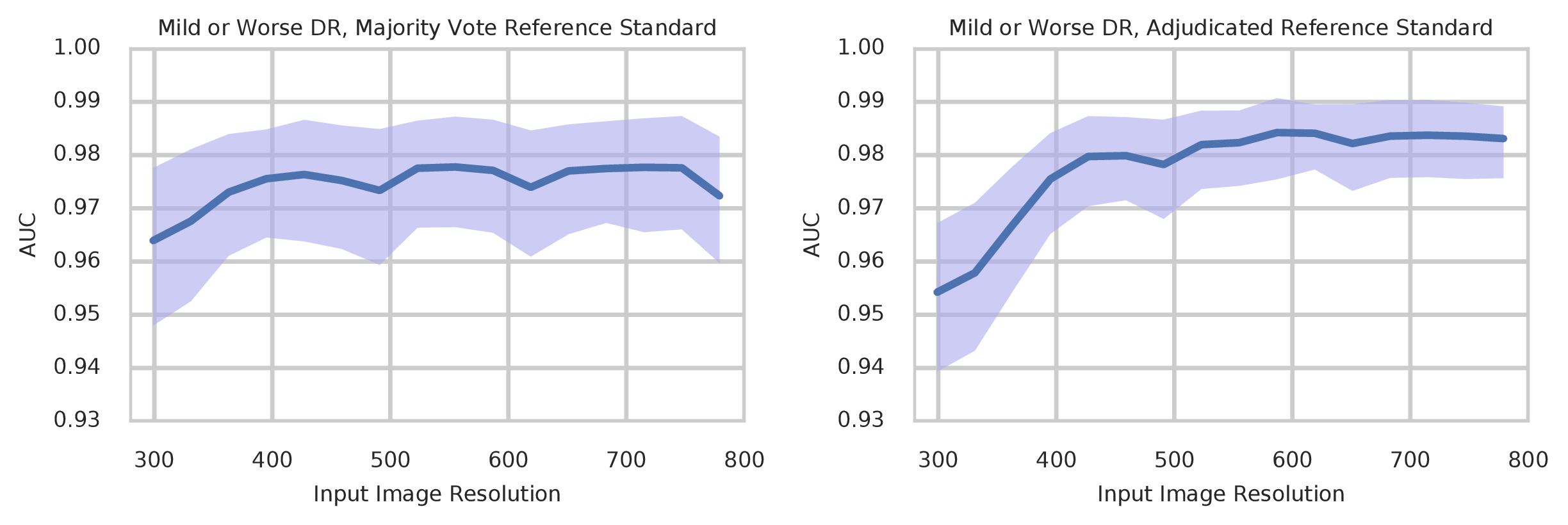}
  \caption{
Image resolution input to model versus area under the curve (AUC) for mild and above DR. Left: Using majority decision of retinal specialists as the reference standard. Right: Using the adjudicated consensus grade of retinal specialists as a reference standard. Shaded areas represent a 95\% confidence interval as measured via bootstrapping.
  }
  \label{fig:res_auc}
\end{figure*}

\begin{figure*}[t]
  \centering
  \includegraphics[width=0.99\linewidth]{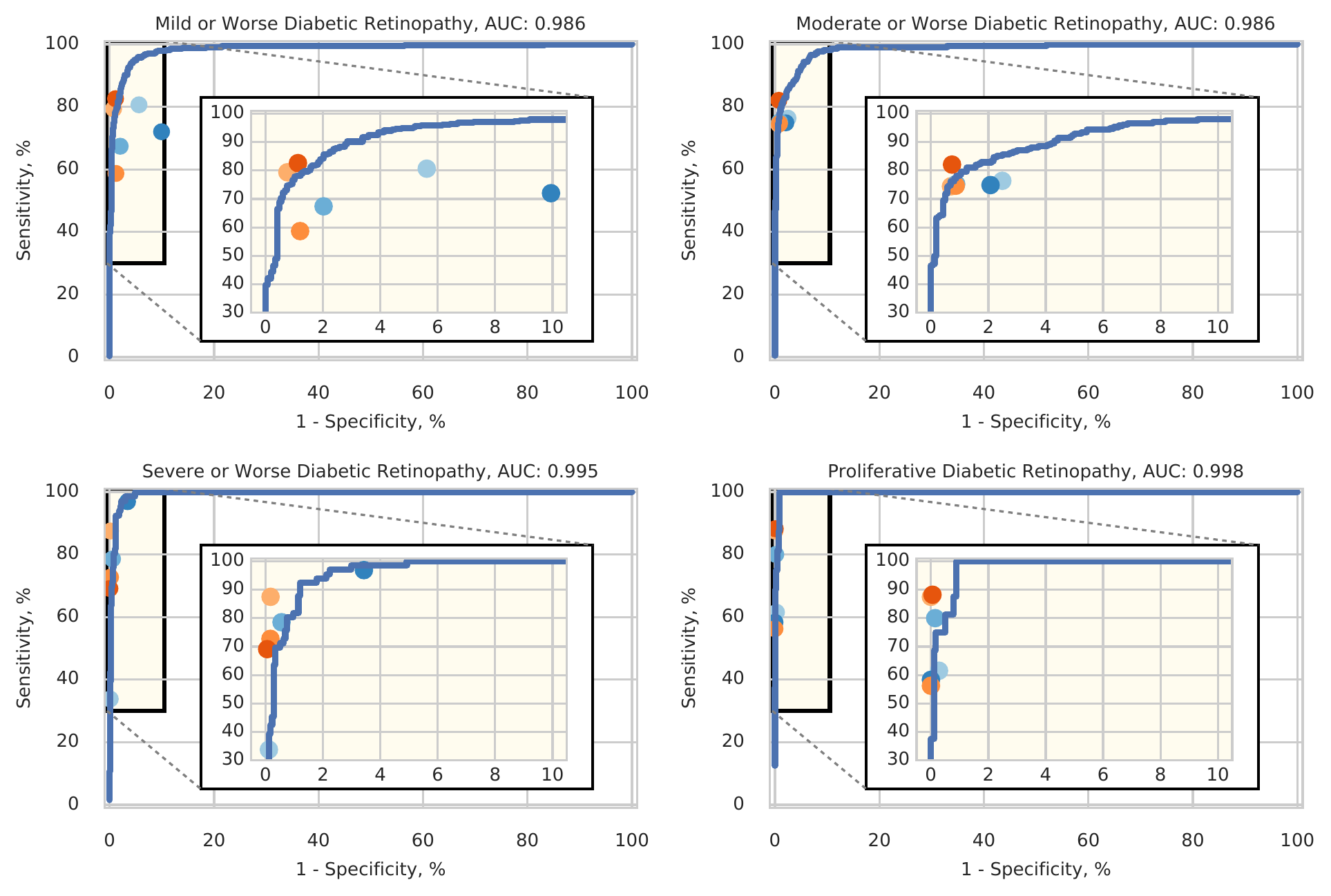}
  \caption{
Comparison of the algorithm, ophthalmologists, and retinal specialists using the adjudicated reference standard at various DR severity thresholds. The algorithm's performance is the blue curve. The 3 retina specialists are represented in shades of orange/red, and the 3 ophthalmologists are in shades of blue. N = 1813 fully gradable images.
  }
  \label{fig:roc_curves}
\end{figure*}

\appendix
\newpage
\FloatBarrier
\setcounter{table}{0}
\renewcommand{\thetable}{S\arabic{table}}%
\setcounter{figure}{0}
\renewcommand{\thefigure}{S\arabic{figure}}%

\begin{table}[t]
  \centering
  \begin{tabular}{|l|l|r|r|r|r|r|}
    \hline
    & & \multicolumn{5}{c|}{\makecell{\textbf{Majority of retinal specialist}\\\textbf{ and ophthalmologist grading}}} \\ \hline
    & & \textbf{No} & \, \textbf{Mild} & \, \textbf{Moderate} \, & \textbf{Severe} & \, \textbf{Proliferative} \\ \hline
    \multirow{ 5}{*}{\makecell{\textbf{Adjudicated}\\\textbf{Consensus}}} & \textbf{No} & \, 1,472 & 3 & 3 & 0 & 0 \\ \cline{2-7}
    & \textbf{Mild} & 65 & 55 & 5 & 0 & 0 \\ \cline{2-7}
    & \textbf{Moderate} & 26 & 6 & 109 & 3 & 0 \\ \cline{2-7}
    & \textbf{Severe} & 0 & 0 & 10 & 40 & 0 \\ \cline{2-7}
    & \textbf{Proliferative} & 0 & 0 & 1 & 2 & 13 \\ \hline
  \end{tabular}
\vspace{5mm}
\newline
\centering
  \begin{tabular}{|l|l|r|r|}
    \hline
    & & \multicolumn{2}{c|}{\makecell{\textbf{Majority of retinal specialist}\\\textbf{and ophthalmologist grading}}} \\ \hline
    & & \, \textbf{Not Referable} & \, \textbf{Referable DME} \\ \hline
    \multirow{ 2}{*}{\makecell{\textbf{Adjudicated}\\\textbf{Consensus}}} & \textbf{Not Referable} & 1,695 & 2 \\ \cline{2-4}
    & \textbf{Referable DME} & 23 & 93 \\ \hline
  \end{tabular}
  \vspace{4mm}
  \caption{
Comparison of combined retina specialist and ophthalmologist grades versus adjudicated grades from retina specialists. Confusion matrix for DR and DME where the grade is determined by either the majority decision of all six retina specialists and ophthalmologists or is the adjudicated consensus of the retinal specialists.
}
  \label{table:comparison_specialist_and_oph}
\end{table}
\newpage

\FloatBarrier

\begin{figure*}[t]
  \centering
  \includegraphics[width=0.98\columnwidth]{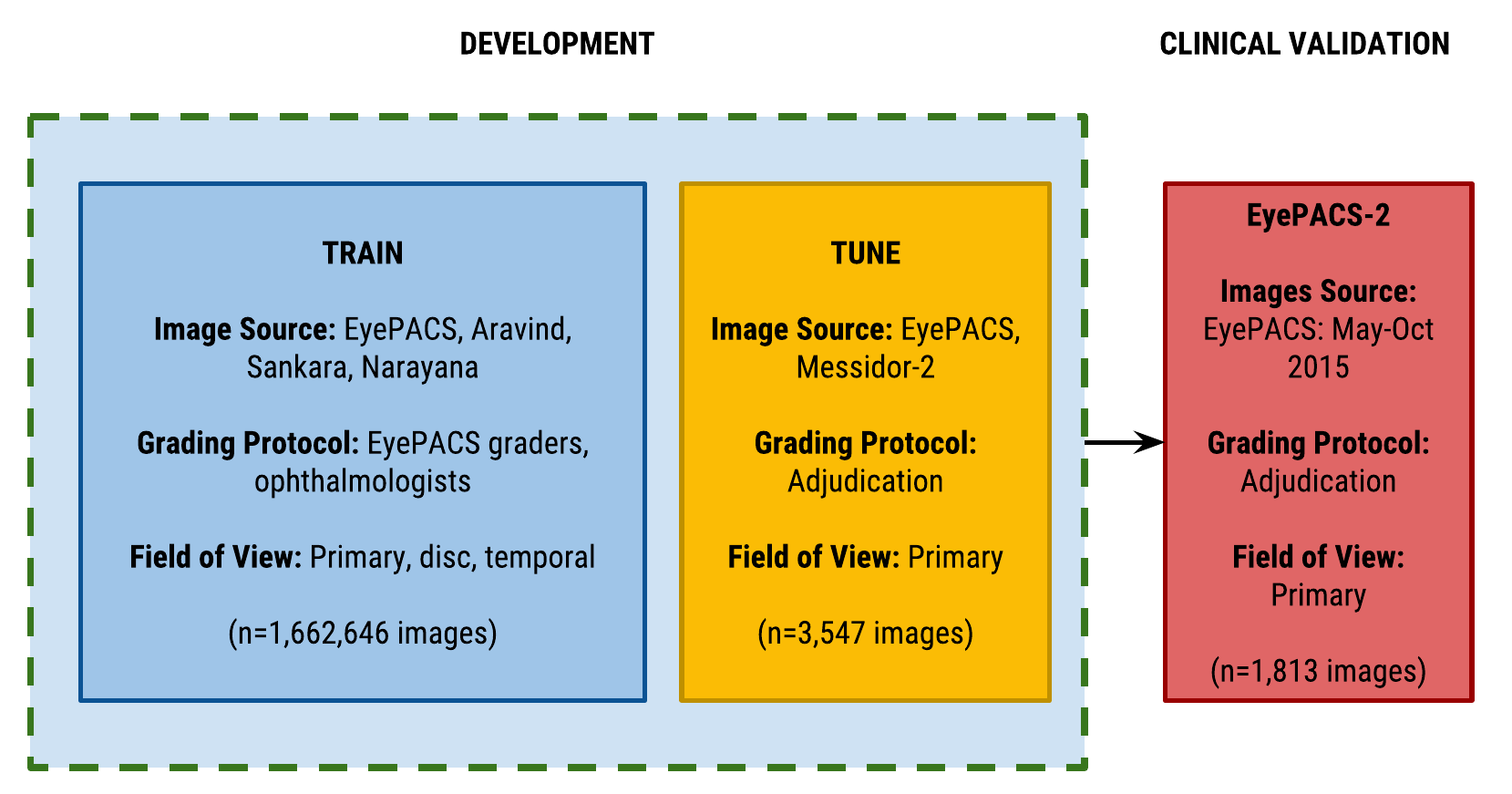}
  \caption{
 Diagram of the different image sources and grading protocols used for development and clinical validation. The algorithm was trained on the development set only, which consisted of a “train” set to train the model parameters and the “tune” set to tune the models performance (hyperparameter tuning and picking checkpoints).
  }
  \label{fig:image_source_protocols}
\end{figure*}

\begin{figure*}[t]
  \centering
  \includegraphics[height=100mm]{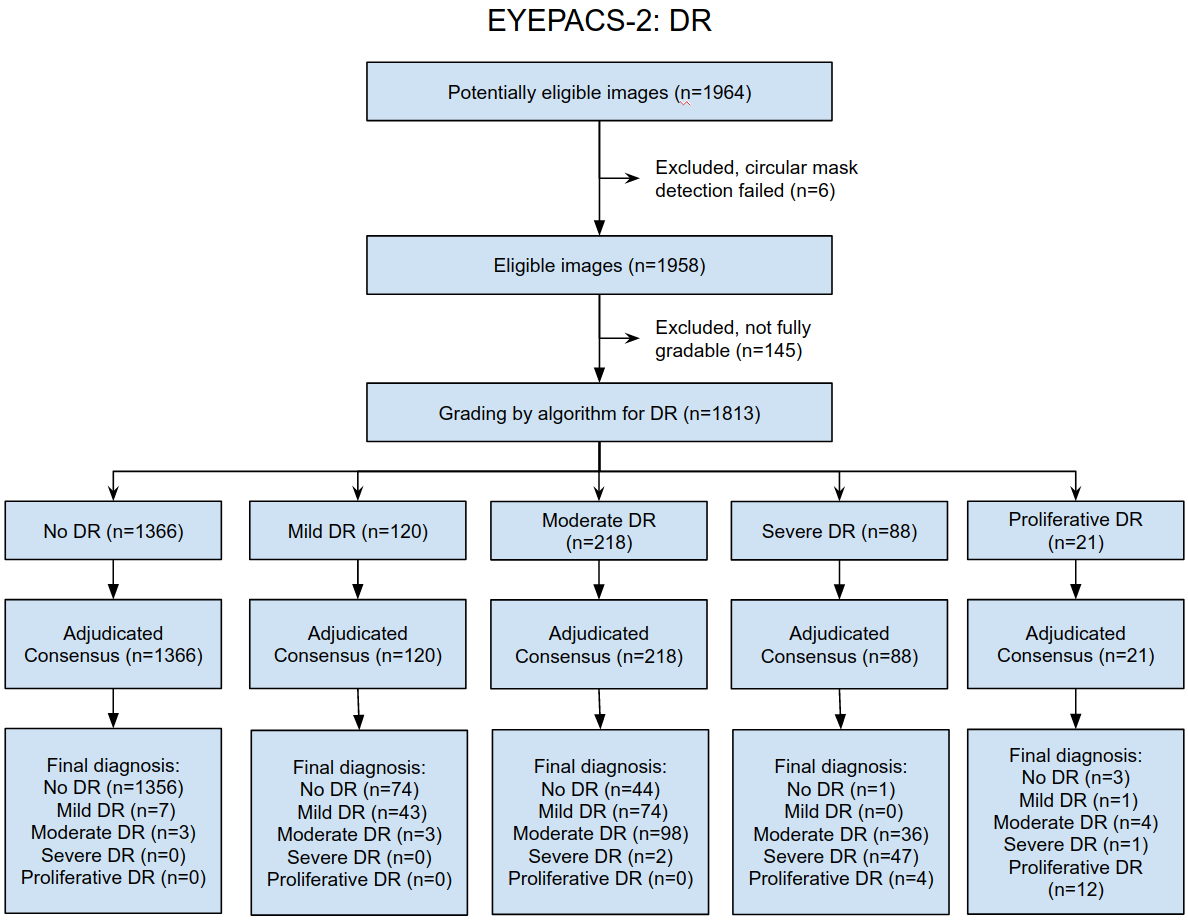}
  \caption{
STARD diagram for DR.
  }
  \label{fig:stard_dr}
\end{figure*}

\begin{figure*}[t]
  \centering
  \includegraphics[height=100mm]{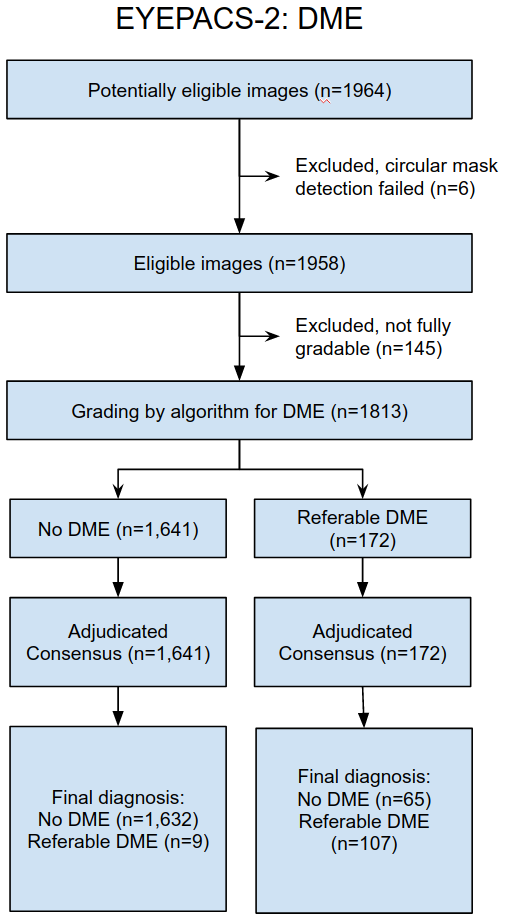}
  \caption{
STARD diagram for DME.
  }
  \label{fig:stard_dme}
\end{figure*}

\end{document}